\documentclass[10pt,conference]{IEEEtran}

\usepackage{float}
\usepackage{xcolor}

\ifCLASSINFOpdf
   \usepackage[pdftex]{graphicx}
   \graphicspath{{figs/}}
   \DeclareGraphicsExtensions{.pdf,.jpeg,.png}
\else
   \usepackage[dvips]{graphicx}
   \graphicspath{{../figs/}}
   \DeclareGraphicsExtensions{.eps}
\fi

\usepackage[cmex10]{amsmath}
\usepackage{amsthm}

\usepackage{comment}

\usepackage{algorithmic}

\usepackage{array}

\ifCLASSOPTIONcompsoc
  \usepackage[caption=false,font=normalsize,labelfont=sf,textfont=sf]{subfig}
\else
  \usepackage[caption=false,font=footnotesize]{subfig}
\fi

\usepackage{booktabs}
\usepackage{url}

\usepackage{multirow}
\IEEEoverridecommandlockouts

\begin{document}

\title{Scene Change Detection Using Multiscale \\ Cascade Residual Convolutional Neural Networks}

%------------------------------------------------------------------------- 
% change the % on next lines to produce the final camera-ready version 
\newif\iffinal
%\finalfalse
\finaltrue
%\newcommand{\cmtid}{71}
%------------------------------------------------------------------------- 
\iffinal
\author{%
    \IEEEauthorblockN{Daniel~F.~S.~Santos$^{\dag}$\thanks{$^{\dag}$These authors contributed equally to this paper.}, Rafael~G.~Pires$^{\dag}$}
    \IEEEauthorblockA{
      Department of Computing\\
      S\~ao Paulo State University\\
      Bauru, Brazil\\
      \{danielfssantos1, rafapires\}@gmail.com}
  \and
	\IEEEauthorblockN{Danilo~Colombo}
	\IEEEauthorblockA{%
	  Cenpes\\
	  Petroleo Brasileiro S.A. - Petrobras\\
	  Rio de Janeiro - RJ, Brazil\\
	  colombo.danilo@petrobras.com.br
  \and
	\IEEEauthorblockN{Jo\~{a}o~P.~Papa}
	\IEEEauthorblockA{%
      Department of Computing\\
      S\~ao Paulo State University\\
      Bauru, Brazil\\
      joao.papa@unesp.br}}}
\else
  \author{Sibgrapi paper ID: \cmtid \\ }
\fi

\maketitle

\begin{abstract}
Scene change detection is an image processing problem related to partitioning pixels of a digital image into foreground and background regions. Mostly, visual knowledge-based computer intelligent systems, like traffic monitoring, video surveillance, and anomaly detection, need to use change detection techniques. Amongst the most prominent detection methods, there are the learning-based ones, which besides sharing similar training and testing protocols, differ from each other in terms of their architecture design strategies. Such architecture design directly impacts on the quality of the detection results, and also in the device resources capacity, like memory. In this work, we propose a novel Multiscale Cascade Residual Convolutional Neural Network that integrates multiscale processing strategy through a Residual Processing Module, with a Segmentation Convolutional Neural Network. Experiments conducted on two different datasets support the effectiveness of the proposed approach, achieving average overall $\boldsymbol{F\text{-}measure}$ results of $\boldsymbol{0.9622}$ and $\boldsymbol{0.9664}$ over Change Detection 2014 and PetrobrasROUTES datasets respectively, besides comprising approximately eight times fewer parameters. Such obtained results place the proposed technique amongst the top four state-of-the-art scene change detection methods.
\end{abstract}


\IEEEpeerreviewmaketitle

\section{Introduction}\label{s.intro}

Scene change detection is a specific kind of image processing task, that involves partitioning the digitalized captured scene into foreground and background pixel regions. Such a processing strategy is frequently used in many visual knowledge-based computer intelligent systems, such as traffic monitoring~\cite{kato2002hmm}, autonomous driving~\cite{dai2019hybridnet}, object and people tracking~\cite{zhou2005real}, action recognition~\cite{feichtenhofer2019slowfast}, video surveillance~\cite{brutzer2011evaluation}, and anomaly detection~\cite{chandola2009anomaly}. Each of those systems presents its challenges for the change detection itself, such as: (a) the shooting environment condition, (b) video capture device quality, and also (c) local computer memory storage capacity.

Concerning some difficulties presented by (a), it can be named a few ones such as shadows, low-light, specular reflections, and blizzard. Regarding (b), it can be noticed problems with the device sensors, mostly due to subtle temperature variations and also issues related to digital noise, mainly generated during analogic to digital signal conversion. Regarding (c), the change detection technique must be adaptable to work in mobile-reduced memory devices such as smartphones, tablets, and drones.

In the last few decades, in an attempt to solve problems (a), (b), and (c), many scene change detection techniques have been developed. They can be classified into two big groups, i.e., the non-learning-based and the learning-based ones. Amongst the non-learning-based group, one can refer to the works of KaewTraKulPong and Bowden~\cite{kaewtrakulpong2002mog}, Zivkovic~\cite{zivkovic2004improvedgmm}, and Varadarajan et al.~\cite{varadarajan2013spatial}, with a strong basis on statistical parametric modeling of the scene changes. Considering the same statistical domain, one can also encounter the works of Bevilaqua et al.~\cite{bevilacqua2005cnm}, and Lanza and Di Stafano~\cite{lanza2011statchange}, that use nonparametric statistics for the scene change modeling. Besides such mentioned techniques, it is possible to find more simple and effective methods, which include SuBSENSE from St-Charles et al.~\cite{st2014subsense}, PWCS from~ St-Charles et al.~\cite{st2015pawcs}, and IUTIS-5 from Bianco et al.~\cite{bianco2017iutis5}.

The second group of change detection techniques includes those methods capable of learning how to differentiate between the foreground and background scene regions, that when properly designed and trained, can easily adapt to difficult change detection scenarios, as demonstrated by the works of Wang et al.~\cite{wang2017cascadecnn}, that use a multistage and multiscale network named Cascade, Babaee et al.~\cite{babaee2018deepbs}, concerning the usage of a multistage convolutional neural network named DeepBS, Santos et al.~\cite{santos2019crcnn}, which use a multistage residual convolutional neural network named CRCNN, Santana et al.~\cite{santana2019novel}, which use siamese-based change detection networks named SEU-Nets, and Lim and Keles~\cite{segnet2018tripletcnns}\textendash\cite{lim2019fgsegnetv2}, that work with autoencoder change detection convolutional neural networks FgSegNet\_M, FgSegNet\_S, and FgSegNet\_v2.

%%%%%%%%%%%%%%%%%% Inserting here the proposed approach figure
\begin{figure*}[!htb]
\centering
  \includegraphics[width=4.5in,height=4.5in,keepaspectratio]{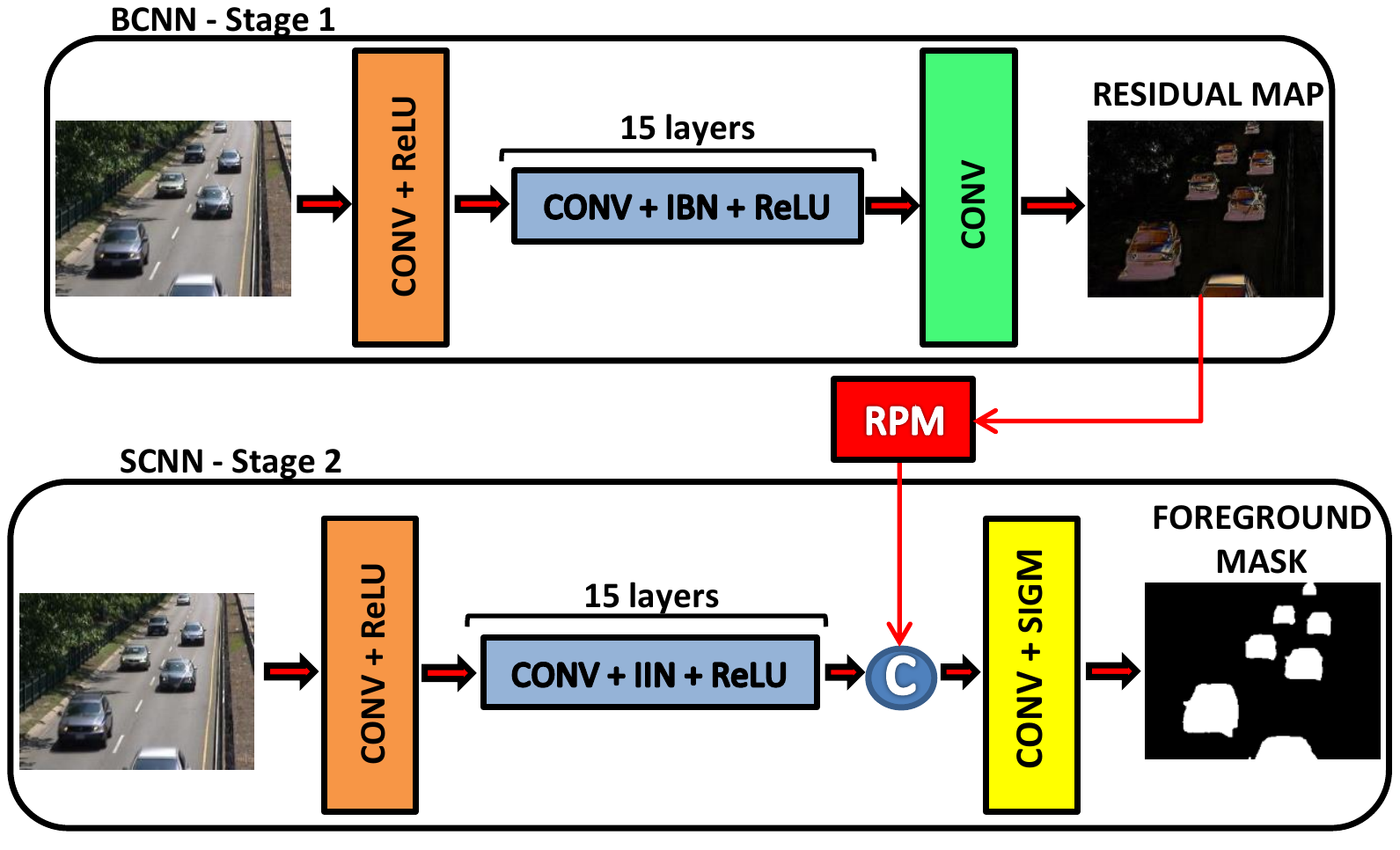}
  \caption{Architecture of the proposed MCRCNN model, where the output of the residual processing module \textbf{RPM} is depth-wise concatenated to the 15th SCNN convolutional layer feature maps. \textbf{CONV} stands for convolutional layers, while \textbf{IBN} and \textbf{IIN} indicate, respectively, the interleaved batch normalization and instance normalization layers, which applied at every three subsequent convolutional layers.}
  \label{f.mcrcnn_arch}
\end{figure*}
 
Although the learning-based methods present state-of-the-art results in the literature when compared against the non-learning-based techniques, they are not yet capable at solving, at the same time, the problems (a), (b), and (c). The CascadeCNN, DeepBS, and CRCNN techniques can be low memory consumptive methods, but at the same time, do not achieve FgSegNets results. On the other hand, in the case of the FgSegNets, better detection results implicate high memory consumption.

In this work, we attempt to improve the effectiveness of the CRCNN method in dealing with problems (a) and (b), trying to maintain the technique already good compromise with the problem (c). In that sense, we propose four modifications to improving the CRCNN method. Such changes include: (i) Multiscale residual map processing (ii) Multistage training using high-level feature aggregation policy (iii) Interleaved and hybrid intralayer feature normalization using batch~\cite{ioffe2015batchnorm} and instance~\cite{ulyanov2016instnorm} normalization strategies, and (iv) color image processing.  

The remaining of this manuscript is divided into Section~\ref{s.proposal}, describing the theoretical basis of the proposal, Section~\ref{s.methodology}, presenting the proposal training and evaluation methodology, Section~\ref{s.experimental_section}, presenting and discussing the quantitative and qualitative obtained results, and Section~\ref{s.conclusion}, showing the conclusion of this work and pointing towards future research directions.
 
\section{Proposed Approach}\label{s.proposal}

In this work, we propose a learning-based scene change detection technique named Multiscale Cascade Residual Convolutional Neural Network (MCRCNN). Such a proposal is based on the work of Zhang et al.~\cite{zhang2017dncnn}, concerning the usage of residual learning and on the work of Santos et al.~\cite{santos2019crcnn}, regarding the usage of a multistage cascaded convolutional neural network for scene change detection. Figure~\ref{f.mcrcnn_arch} summarizes the MCRCNN proposal, which consists of a two-stage deep convolutional neural network composed of $20$ layers and a multiscale Residual Processing Module (RPM).

The first stage of the MCRCNN model consists of learning how to generate the so-called residual map, as described in more detail in Subsection~\ref{s.bcnn_net}. In the second stage, the multiscale processed residual map, as described by Subsection~\ref{s.rpm}, is integrated into the change detection network, whose functionality is described by Subsection~\ref{s.scnn_net}.

\subsection{Background Convolutional Neural Network}\label{s.bcnn_net}

The first change detection stage of the MCRCNN model, named Background Convolutional Neural Network (BCNN), is responsible for generating a foreground highlighted image, such as the vehicles in Figure~\ref{f.mcrcnn_arch}. The BCNN architecture is very similar to the Denoising Convolutional Neural Network (DnCNN) proposed by Zhang et al.~\cite{zhang2017dncnn}. As shown by Figure~\ref{f.mcrcnn_arch}, it starts with a single convolutional layer, shown in orange color, gets deeper with the insertion of $15$ more convolutional layers, represented by the blue-colored rectangle, and ends with a single convolutional layer, shown in green color.

Blue-colored and orange-colored layers in Figure~\ref{f.mcrcnn_arch} are locally activated by Rectified Linear Unity (ReLU) functions~\cite{nair2010relu}, use kernels of size $3\times3$, and output $64$ feature maps each. The green-colored layer is linearly activated, uses kernels of size $3\times3$, and outputs the residual map color image. One particularity of the blue-colored layers is the Interleaved Batch Normalizations (IBNs), which are batch normalization~\cite{ioffe2015batchnorm} operations applied at intervals of three layers, just before the ReLU activation procedure. Such a strategy tries to equally distribute the normalization procedure along the entire network avoiding processing overhead, also diminishing the network memory consumption.

The BCNN training procedure follows the same principles of the CRCNN work~\cite{santos2019crcnn}. It consists of two phases: the first one takes an interval $I = \{S_{1}, S_{2}, ..., S_{m}\}$ of consecutive frames from the video and uses it to calculate the \emph{deterministic background image}, which stands for an image $s$ that represents the median of such an interval\footnote{The same procedure was adopted by Lanza et al.~\cite{bevilacqua2005cnm}. Other alternatives would be using auxiliary non-learning-based segmentation techniques, like performed by Babaee et al.~\cite{babaee2018deepbs}, or even manually selection.}. The second phase consists of minimizing the accumulated\footnote{Minimizing the sum rather than the mean cost value imposes to the optimization an even bigger penalization.} square error between the deterministic background image and the \emph{approximated background image} $b$, which is represented as follows:

\begin{equation}
\label{e.approximated_background}
b = f - BCNN(f; \Theta_{1}),
\end{equation}
where $f$ denotes the input image normalized between $[0, 1]$, $\Theta_{1}$ refers to the BCNN trainable parameters, and $BCNN(\cdot)$ refers to the residual map learned during the training process. In light of that, the BCNN training process aims at minimizing the following equation:

\begin{equation}
\label{e.backcnn_cost}
L_{B}(b, f; \Theta_{1}) = \frac{1}{2}\sum_{i=1}^n||b_{i} - s_{i} ||_F^2,
\end{equation}
where $n$ stands for the number of training samples and $||\cdot||_F^2$ represents the Frobenius norm. Notice that we employed a patch-based methodology, where $b_i$ and $s_i$ denote the $i^{th}$ patch extracted from images $b$ and $s$, respectively.  

\subsection{Residual Processing Module}\label{s.rpm}

\begin{figure}[!ht]
\centering
\includegraphics[width=3.5in]{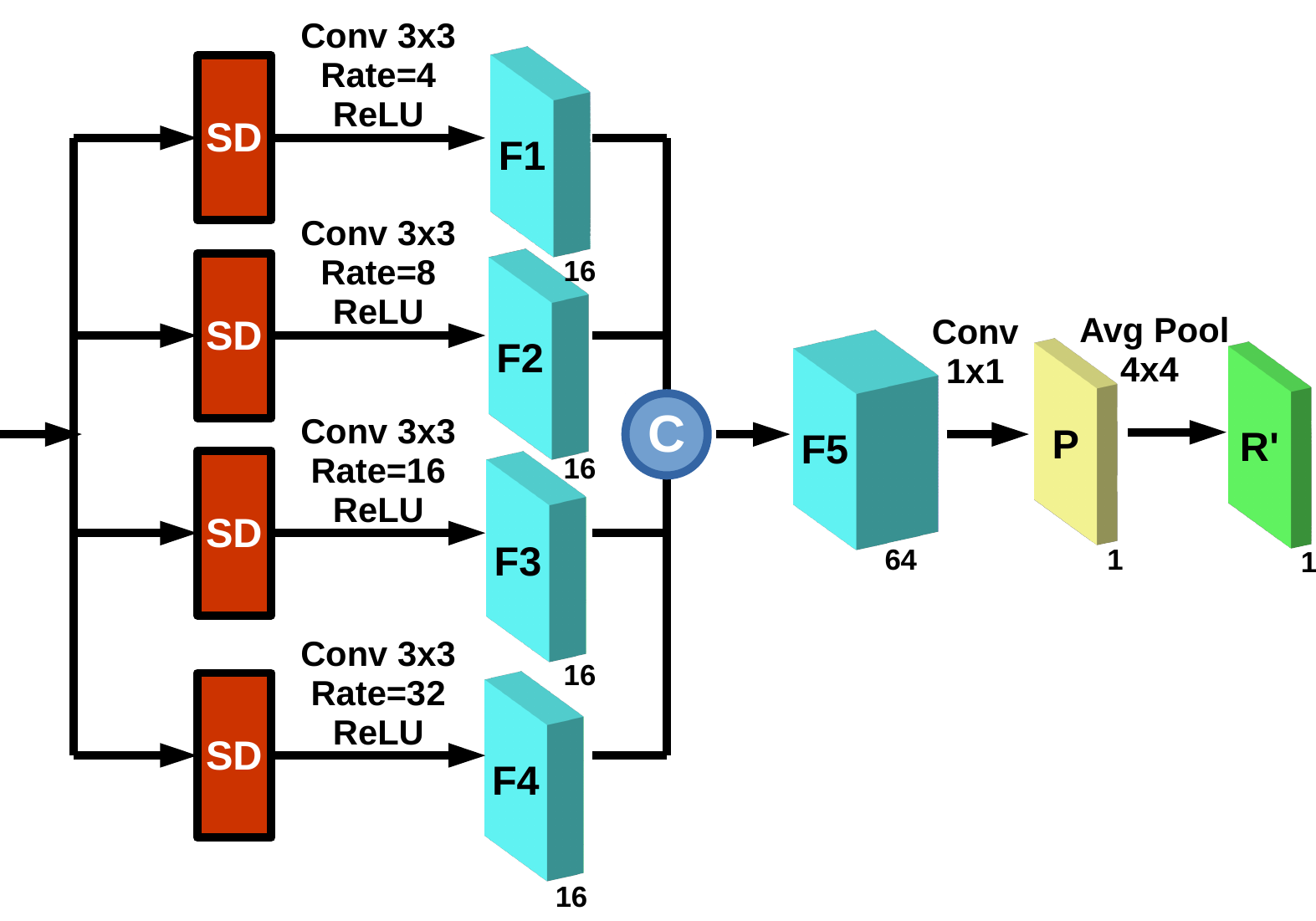}
\caption{Residual Processing Module architecture.}
\label{f.rpm}
\end{figure}

The Residual Processing Module (RPM) design was inspired by the Feature Processing Modules FPM~\cite{segnet2018tripletcnns} and FPM\_M~\cite{lim2019fgsegnetv2}. It serves mainly to improve the BCNN residual map quality treating undesirable spatial coherence problems. As shown by Figure~\ref{f.rpm}, RPM starts by applying, over the residual map, a Spatial Dropout (SD) pre-processing technique, which according to Hinton et al.~\cite{hinton2012dropout} is a very efficient strategy to prevent the network parameters from overspecialization.   

After the SD regularization, the residual map is conducted to the multiscale processing stage, where it is convolved by dilated\footnote{Such a strategy tries to simulate the usage of kernel sizes of respectively $7\times7$, $11\times11$, $18\times18$, and $35\times35$.} filters at rates of $4$, $8$, $16$, and $32$. The dilation results are then activated by ReLU functions, which generate the feature maps \textbf{F1} to \textbf{F4}.

Such generated feature maps are next depth-wise concatenated into \textbf{F5} and convolved by a single $1\times1$ filter, producing the local linear-combined map \textbf{P}. In the last RPM processing step, \textbf{P} is smoothed by an average pooling layer of window size $4\times4$, generating the refined residual map \textbf{R'}, that after has been properly normalized\footnote{The values of the output of the RPM module are normalized between $[0, 1]$ using min-max normalization.}, is used in the second stage of the MCRCNN proposed model.

\subsection{Segmentation Convolutional Neural Network}\label{s.scnn_net}

The second stage of the MCRCNN model is named Segmentation Convolutional Neural Network (SCNN) and it is responsible to generate the probability map identifying, with real values between $[0, 1]$, the image change locations, also called foreground regions. In this multiscale version of the CRCNN proposed by Santos et al.~\cite{santos2019crcnn}, the RPM output (see Subsection~\ref{s.rpm} for more details) is depth-wise concatenated with the 15th SCNN convolutional layer output\footnote{Such output comprehends a set of $64$ feature maps activated by ReLU function.}. The resultant block of $65$ feature maps is then convolved by a single filter of size $3\times3$ and activated by a sigmoid function, been such convolutional process represented in Figure~\ref{f.mcrcnn_arch} by the yellow rectangle.    

The SCNN normalization policy follows the BCNN one, but in such case, the IBNs are substituted\footnote{Since the SCNN optimization consists in using the full-sized images, IN processing adapts better than BN ones.} by Interleaved Instance Normalizations (IINs)~\cite{ulyanov2016instnorm}. The training process follows the work by~\cite{santos2019crcnn}, which aims at minimizing the average binary cross-entropy measured between the network output and the ground-truth binary detection mask. Such an image corresponds to the pre-annotated true foreground regions present in the grayscale input image. Therefore, the SCNN training process aims at minimizing the following equation:

\begin{equation}
\label{e.segcnn_cost}
\begin{aligned}
L_{S}(t, f; \Theta_{2})=-\sum_{i=1}^{k1}\sum_{j=1}^{k2}&[t_{i, j}\log(\hat{t}_{i, j}) \ +\\
       &(1 - t_{i, j})\log(1 - \hat{t}_{i,j})],
\end{aligned}
\end{equation}
where
\begin{equation}
\label{e.segcnn_cost_p1}
\hat{t} = SCNN(f; \Theta_{2}),
\end{equation} 
\\*
notice that $t$ is the ground-truth pre-annotated binary mask, $\Theta_{2}$ stands for the SCNN trainable parameters, $f$ indicates the SCNN input color image, the same BCNN input image, and $k1$ and $k2$ denote the maximum image height and width, respectively.
\section{Methodology}\label{s.methodology}

In this section, we present the methodology used to train and evaluate the proposed MCRCNN model. To simplify the explanation we structured it into Subsection~\ref{ss.datasets}, which presents the relevant information about the datasets used in this work, Subsection~\ref{ss.sub_train_proc}, that describes the proposal training procedures, and Subsection~\ref{ss.sub_evaluation_proc}, which discuss the MCRCNN evaluation protocol.

\subsection{Datasets}
\label{ss.datasets}

\subsubsection{Change Detection Dataset 2014}
\label{sss.change_detection}

The Change Detection Dataset 2014 (CD2014) is a large and freely available dataset of videos collected by Wang et al~\cite{wang2014cdnet} from different realistic, camera-captured, and challenging scenarios. Such a dataset contains $11$ video categories with $4$ to $6$ video sequences each, subdivided into:

\begin{itemize}
\item \textbf{Baseline}: combines mild challenges present in Dynamic Background, Camera Jitter, Intermittent Object Motion, and Shadow categories into four different videos named highway, office, pedestrians, and PETS2006.
 
\item \textbf{Dynamic Background} (Dyn. Bg.): includes scenes from six different videos with so much background motion, e.g., cars and trucks passing in front of a tree shaken. Such video names are boats, canoe, fall, fountain01, fountain02, and overpass.  

\item \textbf{Camera Jitter} (C. Jitter): contains four indoor and outdoor videos captured by unstable video devices, for example vibrating cameras. Those video names are badminton, boulevard, sidewalk, and traffic.   
 
\item \textbf{Intermittent Object Motion} (Int. Obj.): contains six videos with objects that move and then stop for a short while producing ``ghosting" artifacts. Such video names are abandonedBox, parking, sofa, streetLight, tramstop, and winterDriveway.

\item \textbf{Shadow}: six indoor and outdoor videos containing objects surrounding by a strong shadow that could be miss detected as real moving objects. Such video names are backdoor, bungalows, busStation, copyMachine, cubicle, and peopleInShade.

\item \textbf{Thermal}: five videos that have been captured by far-infrared cameras named corridor, diningRoom, lakeSide, library, and park. 

\item \textbf{Bad Weather} (B. Weat.): includes four outdoor videos captured from challenging winter weather conditions, e.g., snowstorms, and fog. Such video names are blizzard, skating, snowFall, and wetSnow. 

\item \textbf{Low Framerate} (L. Frame.): four videos captured varying frame-rates between $0.17$fps and $1$fps. Such video names are port\_0\_17fps, tramCrossroad\_1fps, tunnelExit\_0\_35fps, and turnpike\_0\_5fps.

\item \textbf{PTZ} (PanTZ): four videos captured by pan-tilt-zoom cameras and named continuousPan, IntermittentPan, twoPositionPTZCam, and zoomInZoomOut.

\item \textbf{Turbulence} (Turbul.): four outdoor videos that show air turbulence caused by rising heat. Which are named turbulence0, turbulence1, turbulence2, and turbulence3.   
\end{itemize}

\subsubsection{PetrobrasROUTES}
\label{sss.petrobras}

The PetrobrasROUTES is a private dataset which consists of $281$ high-resolution color images collected from an indoor Petrobras\footnote{Petrobras is a publicly-held company on an integrated basis and specialized in the oil, natural gas, and energy industry~\cite{Petrobrasdataset}.} workspace. The main challenge of such a dataset regards the detection of objects obstructing escape routes.

\subsection{Training procedure}
\label{ss.sub_train_proc}

The training procedure methodology follows basically the same protocols of~\cite{santos2019crcnn}, where for the CD2014 dataset consist of:

\begin{enumerate}
\item to select $300$ color images\footnote{We used the same set of training images from~\cite{lim2019fgsegnetv2} to train the proposed MCRCNN model.} and their $300$ correspondent binary images, which were ground-truth manually annotated.
\item to calculate the deterministic background over the first $100$ images.
\item to train the BCNN network using batches of randomly extracted patches of size $40 \times 40$, like in~\cite{zhang2017dncnn}, from the input and output background images to minimize the cost of Equation~\eqref{e.backcnn_cost}. The patches were augmented using geometric transformations, such as rotation and reflection.
\item to freeze all BCNN network trainable parameters and just train the second MCRCNN part, the SCNN network, and also the RPM module using the full-sized images to minimize the cost of Equation~\eqref{e.segcnn_cost}.
\end{enumerate}

For the PetrobrasROUTES the training procedure consists in:

\begin{enumerate}
\item to select $51$ color images and their $51$ correspondent binary images, which were ground-truth manually annotated.
\item to manually select one of the $51$ color images to be the deterministic background.
\item to follow the same steps 3) and 4) from the CD2014 dataset training protocol.
\end{enumerate}

The BCNN, RPM, and SCNN parameters were trained using the Adam method~\cite{kingma2014adam} by a maximum of $100$ epochs\footnote{Depending on the training video sequence, convergence can be achieved in less than $100$ epochs.}, with $500$ gradient updates per epoch, using a learning rate\footnote{The initial value is reduced by a factor of $0.1$ every time the loss function hits a plateau.} of $0.001$ and batches of size $128$ for the BCNN training process. We trained the MCRCNN parameters with $80\%$ of the input images and used the remaining $20\%$ to evaluate the convergence of the training process.  

%%%%%%%%%%%%%%%%%%%%%%%%%%%%% Table Overall F-measures CD2014
\begin{table*}[hbt!]
\centering
\renewcommand\arraystretch{1.5}
\setlength{\tabcolsep}{.58em}
\caption{Comparison of F-measure results of 11 categories from CD2014 dataset}
\scalebox{1.05}{
\begin{tabular}{crrrrrrrrrrrr}
\toprule 
Methods & Baseline & C.Jitter & B.Weat & Dyn.Bg. & Int.Obj. & L.Frame. & N.Videos & PanTZ   & Shadow & Thermal & Turbul. &  Overall  \\ 
\midrule
FgSegNet\_{v2} \cite{lim2019fgsegnetv2} & \textbf{0.9980}	& \textbf{0.9961}	 & 0.9900	 & \textbf{0.9950}	& 0.9939       & \textbf{0.9579}     & 0.9816	 & \textbf{0.9936}	& 0.9966	& 0.9942    & \textbf{0.9815}	& \textbf{0.9890} \\
FgSegNet\_S \cite{segnet2018tripletcnns} & \textbf{0.9980}   & 0.9951      & \textbf{0.9902}    & 0.9902   & \textbf{0.9942}       & 0.9511      & \textbf{0.9837}     & 0.9837       & \textbf{0.9967}     & \textbf{0.9945}        & 0.9796        & 0.9878 \\
FgSegNet\_M \cite{segnet2018tripletcnns} & 0.9975   & 0.9945      & 0.9838    & 0.9838   & 0.9933       & 0.9558      & 0.9779     & 0.9779      & 0.9954     & 0.9923        & 0.9776        & 0.9865 \\
MCRCNN	    & 0.9938   & 0.9889	     & 0.9632	 & 0.9811	& 0.9893	   & 0.8619	    & 0.9428	 & 0.9344	     & 0.9906    & 0.9765    & 0.9635	   & 0.9622 \\
CRCNN \cite{santos2019crcnn}      & 0.9919   & 0.9799      & 0.9569    & 0.9687   & 0.9755       & 0.8498      & 0.9388     & 0.8967 & 0.9852     & 0.9818        & 0.9637          & 0.9535 \\
Cascade \cite{wang2017cascadecnn}     & 0.9786   & 0.9758      & 0.9451    & 0.9451   & 0.8505       & 0.8804      & 0.8926     & 0.8926 & 0.9593 & 0.8958  & 0.9215  & 0.9272 \\
DeepBS \cite{babaee2018deepbs}      & 0.9580   & 0.8990      & 0.8647    & 0.8647   & 0.6097       & 0.5900      & 0.6359     & 0.6359 & 0.9304 & 0.7583  & 0.8993  & 0.7593 \\
IUTIS-5 \cite{bianco2017iutis5}     & 0.9567   & 0.8332      & 0.8289    & 0.8289   & 0.7296       & 0.7911      & 0.5132     & 0.5132 & 0.9084 & 0.8303  & 0.8507  & 0.7820 \\
PAWCS \cite{st2015pawcs}  & 0.9397   & 0.8137      & 0.8059    & 0.8059   & 0.7764       & 0.6433      & 0.4171     & 0.4171 & 0.8934 & 0.8324  & 0.7667  & 0.7477 \\
SuBSENSE \cite{st2014subsense}    & 0.9503   & 0.8152      & 0.8594    & 0.8594   & 0.6569       & 0.6594      & 0.4918     & 0.4918 & 0.8986 & 0.8171  & 0.8423  & 0.7453 \\ \bottomrule
\end{tabular}}
\label{t.fmeasures}
\end{table*}

\subsection{Evaluation procedure}
\label{ss.sub_evaluation_proc} 

The evaluation process consists in to apply the trained MCRCNN model over each video test image following the protocol:

\begin{itemize}
\item \textbf{Deep Segmentation}: first forward propagating the test images through the trained BCNN model, generating the residual image counterpart, and through the trained SCNN model. Before the last SCNN convolution, we concatenate the residual image to the 15th SCNN convolutional layer outputs. Later, we binarized\footnote{In the majority of the experiments, the best threshold value was $0.7$, except for the categories B. Weat, Dyn. Bg., Int. Obj., and N. Videos, which used values of respectively $0.8$, $0.9$, $0.6$, and $0.9$.} the SCNN probabilistic output.

\item \textbf{Misclassification Rate}: in such a step, we calculated the number of correct and incorrect detections encoded by the True Positives (TPs), i.e, the number of pixels correctly classified as foreground, the True Negatives (TNs), i.e., the number of pixels correctly classified as background, the False Positives (FPs), i.e., the number of background pixels incorrectly classified as foreground, and the False Negatives (FNs), i.e., the number of foreground pixels incorrectly classified as background.     

\item \textbf{Detection Measurements}: in such a step, (TPs), (TNs), (FPs), and (FNs) are combined into four different measures used to evaluate the robustness of the proposed MCRCNN model. Those measures are computed as follows: 

\begin{equation}\label{equ.precision}
Precision = \frac{TP}{TP+FP},
\end{equation}
\\*
\begin{equation}\label{equ.recall}
Recall = \frac{TP}{TP+FN},
\end{equation}
\\*
\begin{equation}\label{equ.fmeasure}
F-measure = 2.0 \times \frac{Recall \times Precision}{Recall + Precision},
\end{equation}
\\*
\noindent and
\\*
\begin{equation}\label{equ.pwc}
PWC = 100.0 \times \frac{FN + FP}{TP + FP + FN + TN} \\[15pt]
\end{equation}
\end{itemize}
where $PWC$ denotes the percentage of wrong classifications.
\section{Experimental Results}\label{s.experimental_section}

In this section, we present the results of the proposed MCRCNN method regarding the comparison against the non-learning-based change detection techniques, IUTIS-5~\cite{bianco2017iutis5}, PAWCS~\cite{st2015pawcs}, SuBSENSE~\cite{st2014subsense}, and the learning-based ones FgSegNet\_v2~\cite{lim2019fgsegnetv2}, FgSegNet\_S~\cite{segnet2018tripletcnns}, FgSegNet\_M~\cite{segnet2018tripletcnns}, Cascade~\cite{wang2017cascadecnn}, DeepBS~\cite{babaee2018deepbs}, and CRCNN~\cite{santos2019crcnn}.

For the sake of clarity, the discussion is subdivided into Subsection~\ref{ss.cd_dataset_res}, which presents the quantitative and qualitative results related to the CD2014 dataset, and Subsection~\ref{ss.petro_dataset_res}, which presents the results regarding PetrobrasROUTES dataset.   

\subsection{CD2014 Dataset Results}\label{ss.cd_dataset_res}

According to Table~\ref{t.fmeasures}, the MCRCNN proposal, in comparison against SuBSENSE, PAWCS, and IUTIS-5 techniques, shows average overall $F\text{-}measure$ improvements of $0.2169$, $0.2145$, and $0.1802$, respectively. In the worst-case scenario, considering the comparison against the learning-based techniques, MCRCNN average overall $F\text{-}measure$ results were lower than FgSegNet\_v2, FgSegNet\_S, and FgSegNet\_M by respectively $0.0268$, $0.0256$, and $0.0243$. Table~\ref{t.fmeasures} also shows that MCRCNN average overall $F\text{-}measure$ results overcome DeepBS, Cascade, and CRCNN ones. In such cases, the results were improved by respectively $0.2029$, $0.035$, and $0.0090$, respectively. 
 
Analyzing Table~\ref{t.overall_results}, one can see that the proposed technique, in the best-case scenario, achieved improvements in $Precision$, $Recall$, and $PWC$ of respectively $0.2196$, $0.1969$, and $1.8883$, regarding the comparisons against SuBSENSE and DeepBS techniques. Table~\ref{t.overall_results} also shows that even so MCRCNN was not capable to overcome the FgSegnets, it gets close $Precision$ results of $0.0046$ and $0.0053$ in comparisons against FgSegNet\_S and FgSegNet\_M, respectively.

%%%%%%%%%%%%%%%%%%%%%%%%%%%%% Table Overall Precision, Recall, and PWC CD2014
\begin{table}[hbt!]
\renewcommand{\arraystretch}{1.5}
\centering
\caption{Comparison of precision, recall and PWC overall results from CD2014 dataset.}
\scalebox{1.07}{
\begin{tabular}{cccc}
\hline
Methods     & Avg. Precision  & Avg. Recall     & Avg. PWC        \\ \hline
FgSegNet\_v2 \cite{lim2019fgsegnetv2} & \textbf{0.9823} & 0.9891 & \textbf{0.0402} \\
FgSegNet\_S \cite{segnet2018tripletcnns} & 0.9751 & \textbf{0.9896} & 0.0461 \\
FgSegNet\_M \cite{segnet2018tripletcnns} & 0.9758 & 0.9836 & 0.0559 \\
MCRCNN      &           0.9705      &         0.9514          &     0.1037            \\
CRCNN \cite{santos2019crcnn}       & 0.9604 & 0.9602 & 0.1348 \\
Cascade \cite{wang2017cascadecnn}     & 0.8997          & 0.9506          & 0.4052          \\
DeepBS \cite{babaee2018deepbs}      & 0.8332          & 0.7545          & 1.9920          \\
IUTIS-5 \cite{bianco2017iutis5}     & 0.8087          & 0.7849          & 1.1986          \\
PAWCS \cite{st2015pawcs}       & 0.7857          & 0.7718          & 1.1992          \\
SuBSENSE \cite{st2014subsense}    & 0.7509          & 0.8124          & 1.6780          \\ \hline
\end{tabular}}
\label{t.overall_results}
\end{table}

It is worth noting that even so the FgSegNets quantitative results, presented by Tables~\ref{t.fmeasures}~and~\ref{t.overall_results}, overcome the MCRCNN method, our proposal network architecture is much more compact. It has a total of $1,116,618$ parameters, while the top two ranked techniques, i.e., FgSegNet\_v2 and FgSegNet\_S, comprise an amount of $9,225,161$ and $7,622,465$ parameters, respectively. Besides, even considering the RPM module size, the MCRCNN almost preserves the same CRCNN size of $1,112,720$ parameters.

According to Figure~\ref{f.cd2014_segmentation}, when comparing the MCRCNN foreground detection masks in row (d) with the CRCNN masks in row (e), it can be noticed that MCRCNN exhibit more problems related to false negatives, been those problems more pronounced in Bad Weather and Shadow category scenes. The first one regards the incomplete detection of the truck body, and the second one concerns the middle person foot and the people heads. Such observations corroborate with the average overall MCRCNN $Recall$ results presented by Table~\ref{t.overall_petro_results}. 

According to Figure~\ref{f.cd2014_segmentation}, the images from row (f) show that the Cascade technique also has some difficulties to detect changes in the Bad Weather and Shadow categories. It presents even worst false negative issues, as it can be seen by the barely detected truck body in the Bad Weather scene, and by the undetected person's head in the Shadow scene. Also, regarding the Shadow category, different from MCRCNN, CRCNN, and FgSegNet\_v2 techniques, Cascade was not able to avoid the false positive shadow regions.    

Besides the foreground masks, row (b) of Figure~\ref{f.cd2014_segmentation} shows us the BCNN normalized residual map. As it can be noticed, the miss detected foreground regions are pretty much related to the dark residual map regions. In such cases, we argue that the RPM dilation processing strategy was not capable of properly fill those map holes, which could contribute to the SCNN paying less attention to such regions during its training procedure. 

\subsection{PetrobrasROUTES Dataset Results}\label{ss.petro_dataset_res}

Considering the experiments conducted over the PetrobrasROUTES dataset, Table~\ref{t.overall_petro_results} shows that the MCRCNN results overcome learning-based state-of-the-art change detection techniques like FgSegNet\_v2, FgSegNet\_S, and CRCNN in terms of at least three of the four used detection measurements.

%%%%%%%%%%%%%%%%%% Tabela Overall F-measure, Precision, Recall, and PWC PetrobrasRoutes
\begin{table}[hbt!]
\renewcommand{\arraystretch}{1.5}
\centering
\caption{Comparison of precision, recall and PWC overall results from PetrobrasROUTES dataset.}
\scalebox{1.08}{
\begin{tabular}{ccccc}
\hline
Methods      & F-measure & Precision & Recall & PWC    \\ \hline
FgSegNet\_v2 \cite{lim2019fgsegnetv2} & 0.9095    & 0.9672    & 0.8583 & 0.5831 \\
FgSegNet\_S \cite{segnet2018tripletcnns}  & 0.9221    & \textbf{0.9770}    & 0.8732 & 0.4287 \\
MCRCNN       & \textbf{0.9664}    & 0.9661    & \textbf{0.9667} & 0.2296 \\
CRCNN \cite{santos2019crcnn}        & 0.9619    & 0.9611    & 0.9627 & \textbf{0.2218} \\ \hline
\end{tabular}}
\label{t.overall_petro_results}
\end{table}

According to Table~\ref{t.overall_petro_results}, in the best case scenario, regarding the comparison against the FgSegNet\_v2 technique, the MCRCNN method exhibit improvements of $0.0524$, $0.1084$, and $0.3535$ in terms of $F\text{-}measure$, $Recall$, and $PWC$ measurements, respectively. In the worst case scenarios, the MCRCNN comparisons against FgSegNet\_S and CRCNN exhibit worsen results of $0.0109$ and $0.0078$ in terms of respectively $Precision$ and $PWC$ measurements.

\begin{figure}[htb!]
\setlength{\tabcolsep}{2pt} % Default value: 6pt
\renewcommand{\arraystretch}{1.5} % Default value: 1
%%%%%%%%%%%%%%%%%% Frame, Ground-Truth block %%%%%%%%%%%%%%%%%%%%%%
\vspace*{-.3cm}
\centerline{
\begin{tabular}{cr}
   {\small Platform}\\
   \includegraphics[width=8cm, height=3.5cm]{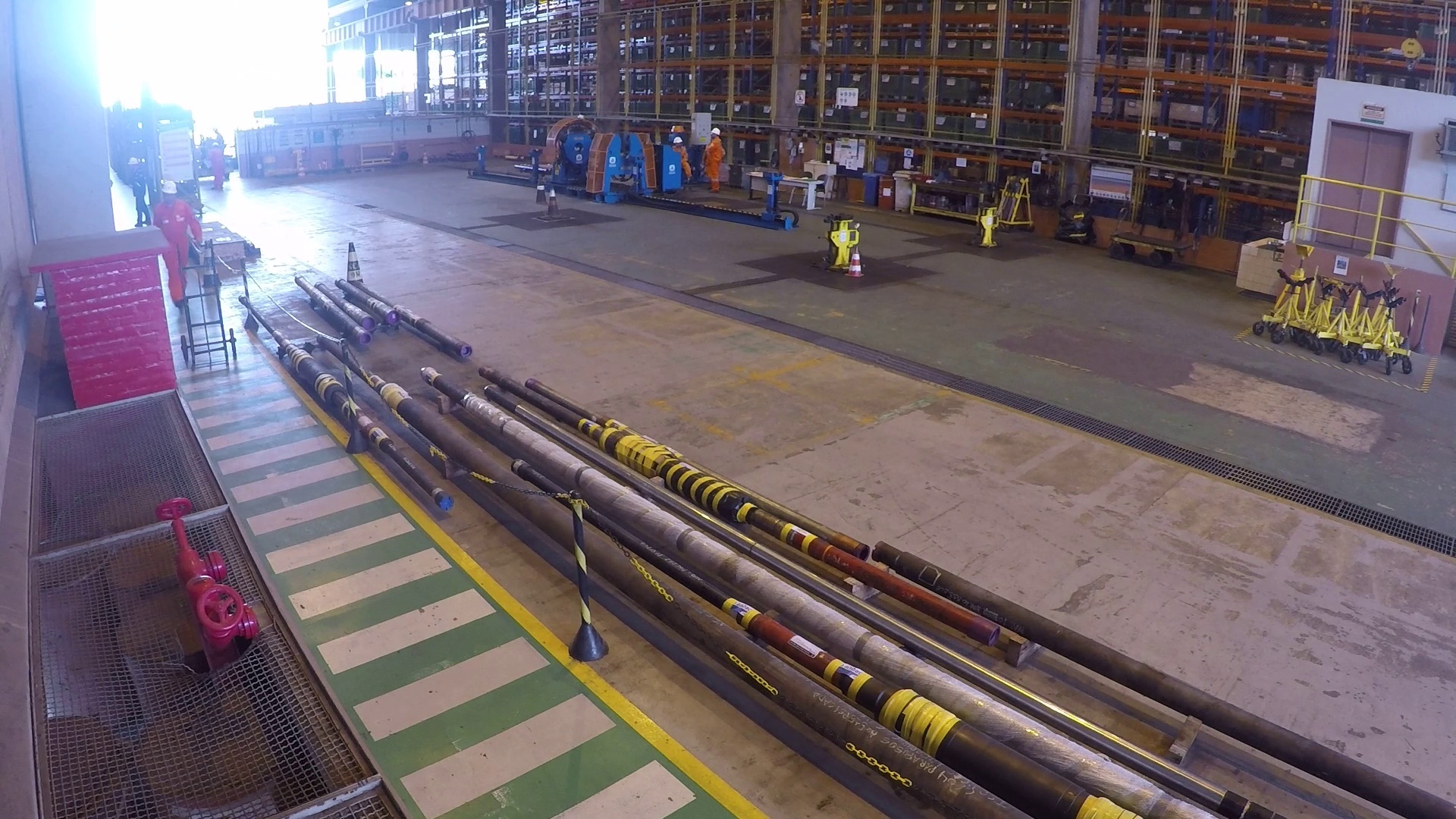}\\
   (a)\\
\end{tabular}}
%%%%%%%%%%%%%%%%%% Ground-Truth block %%%%%%%%%%%%%%%%%%%%%%%
%\hspace*{.05cm}
\centerline{
\begin{tabular}{cr}
   \includegraphics[width=8cm, height=3.5cm]{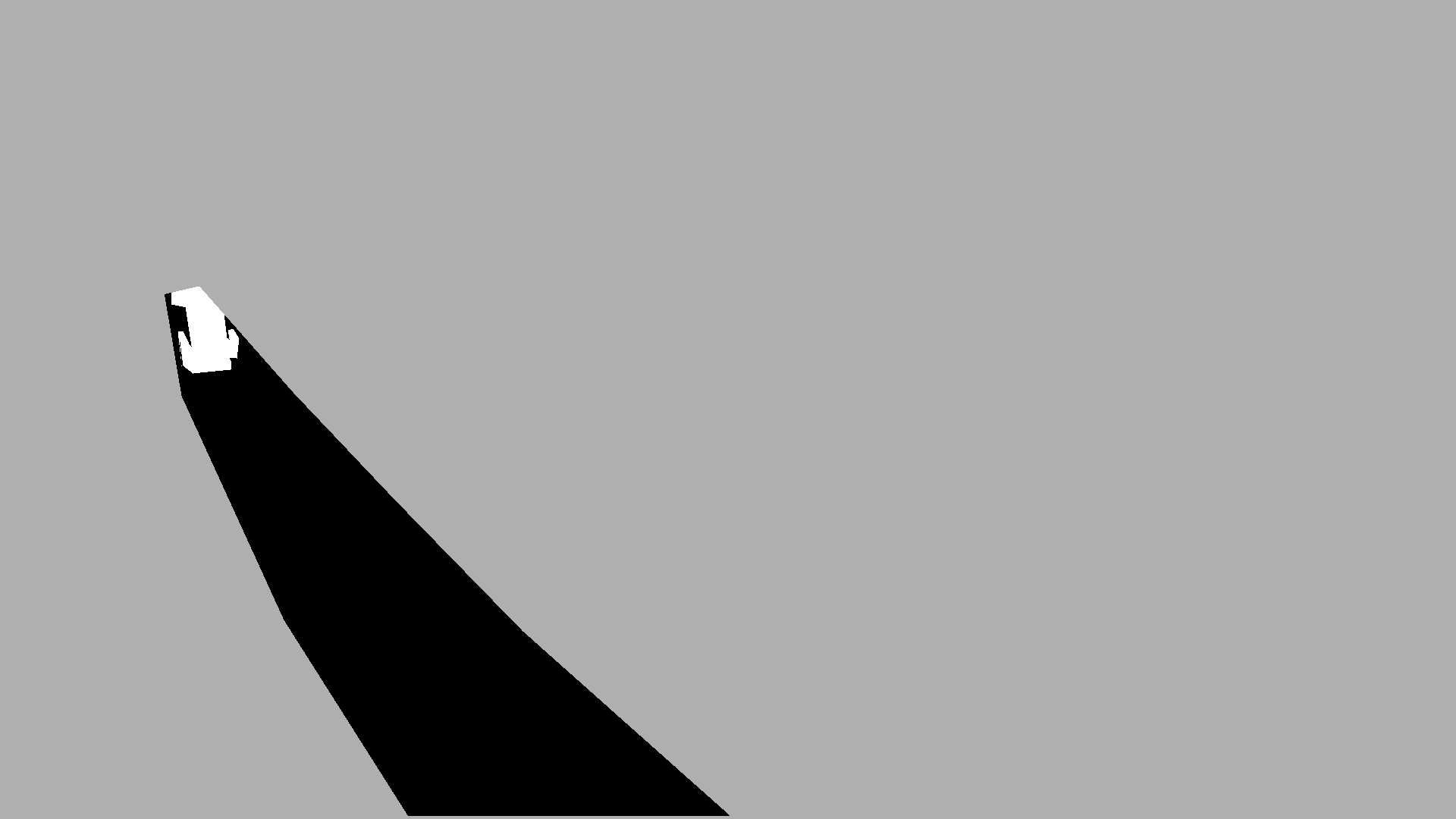}\\
   (b)\\
\end{tabular}}
%%%%%%%%%%%%%%%%%% MCRCNN Block %%%%%%%%%%%%%%%%%%%%%%
\hspace*{.05cm}
\centerline{
\begin{tabular}{cr}
      \includegraphics[width=8cm, height=3.5cm]{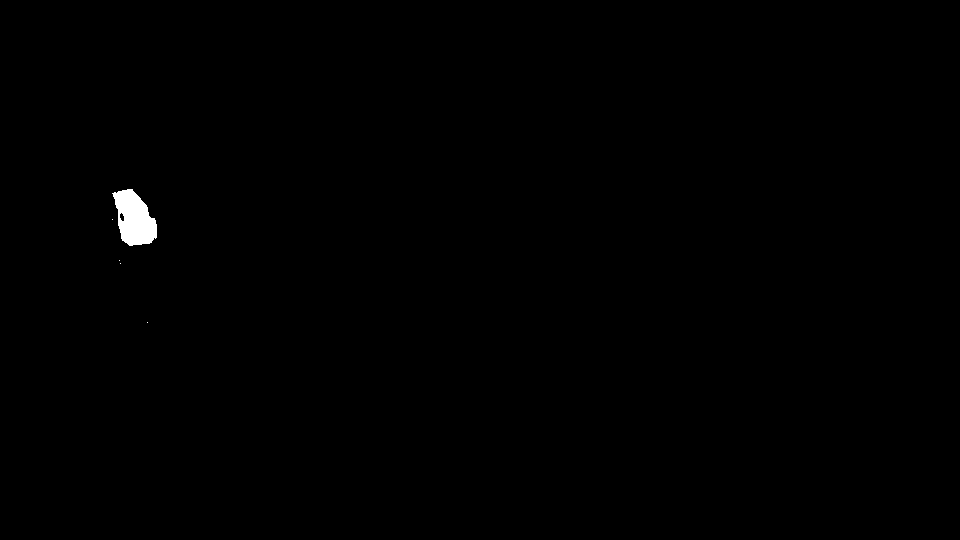}\\
      (c)\\
\end{tabular}}
%%%%%%%%%%%%%%%%%% CRCNN Block %%%%%%%%%%%%%%%%%%%%%%
\hspace*{.05cm}
\centerline{
\begin{tabular}{cr}
      \includegraphics[width=8cm, height=3.5cm]{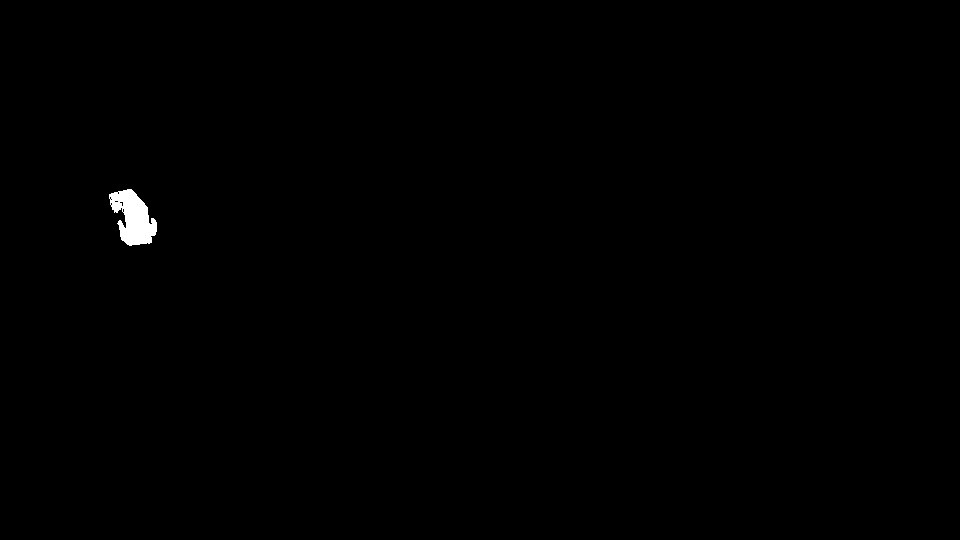}\\
      (d)\\
\end{tabular}}
%%%%%%%%%%%%%%%%%% Segnet_v2 Block %%%%%%%%%%%%%%%%%%%%%%
\hspace*{.05cm}
\centerline{
\begin{tabular}{cr}
      \includegraphics[width=8cm, height=3.5cm]{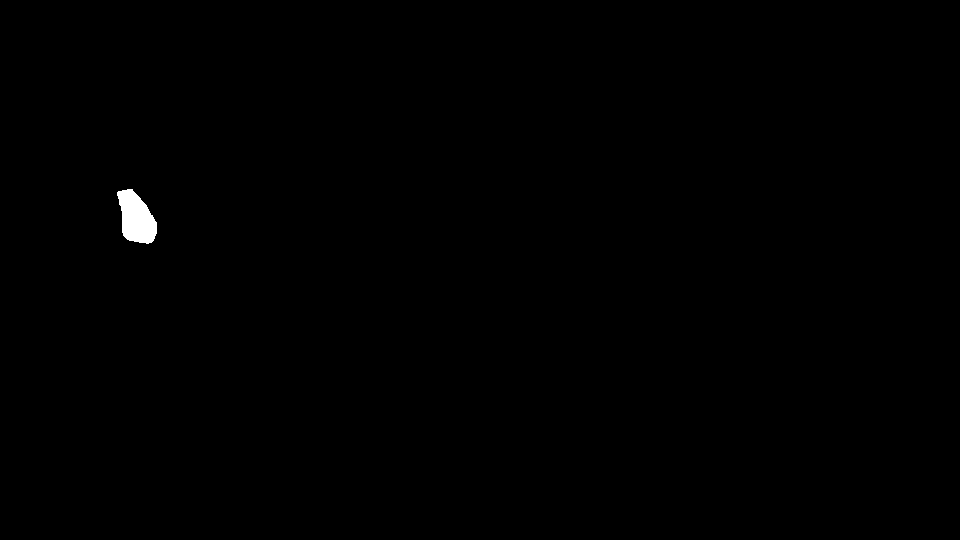}\\
      (e)\\
\end{tabular}}
\centering
\caption{Qualitative results considering an obstructed route video scene from PetrobrasROUTES dataset: (a) input RGB frame, (b) ground-truth detection mask, and results concerning (c) MCRCNN, (d) CRCNN and (e) FgSegNet\_v2 techniques.}
\label{f.petro_segmentation}
\end{figure}

Concerning the detection quality analysis, Figure~\ref{f.petro_segmentation}(c) shows that MCRCNN was able to produce a much more precise foreground object detection mask in comparison against FgSegNet\_v2, whose results were severed affected by false negatives, as shown by Figure~\ref{f.petro_segmentation}(e). On the other hand, even so in Figure~\ref{f.petro_segmentation}(c) most of the object shape was recovered, in comparison against Figure~\ref{f.petro_segmentation}(d), which shows the CRCNN results, and against Figure~\ref{f.petro_segmentation}(b), which shows the reference ground-truth mask, the MCRCNN technique presents more false positive areas around the detected foreground object.        

%%%%%%%%%%%%%%%%%%%%%%%CD2014 figures%%%%%%%%%%%%%%%%%%%%%%%%%%%%%%

\begin{figure}[htb!]
\setlength{\tabcolsep}{2pt} % Default value: 6pt
\renewcommand{\arraystretch}{0.8} % Default value: 1
%%%%%%%%%%%%%%%%%% Frame block %%%%%%%%%%%%%%%%%%%%%%
\hspace*{.05cm}
\centerline{
\begin{tabular}{ccc}
{\small Bad Weather}    &   {\small Low Framerate}  &   {\small Shadow}\\
   \includegraphics[width=2.8cm, height=2.8cm]{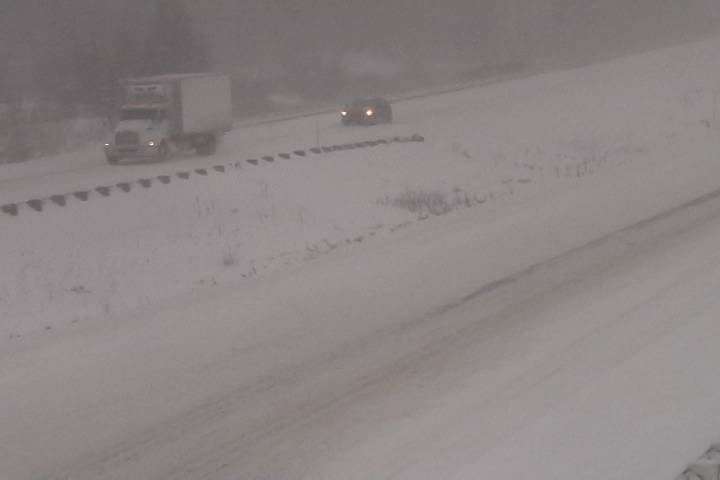}&
      \includegraphics[width=2.8cm, height=2.8cm]{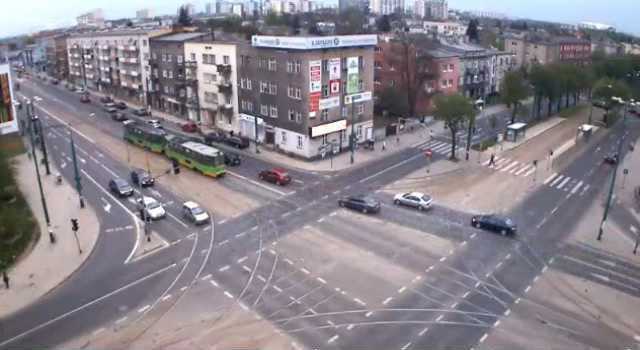}&
      \includegraphics[width=2.8cm, height=2.8cm]{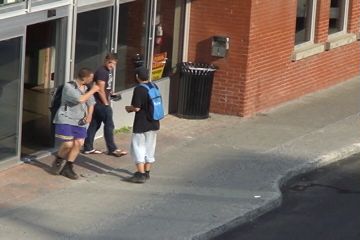}\\
      & (a) & \\
\end{tabular}}
%%%%%%%%%%%%%%%%%% Residual block %%%%%%%%%%%%%%%%%%%%%%
\hspace*{.05cm}
\centerline{
\begin{tabular}{rccc}
      \includegraphics[width=2.8cm, height=2.8cm]{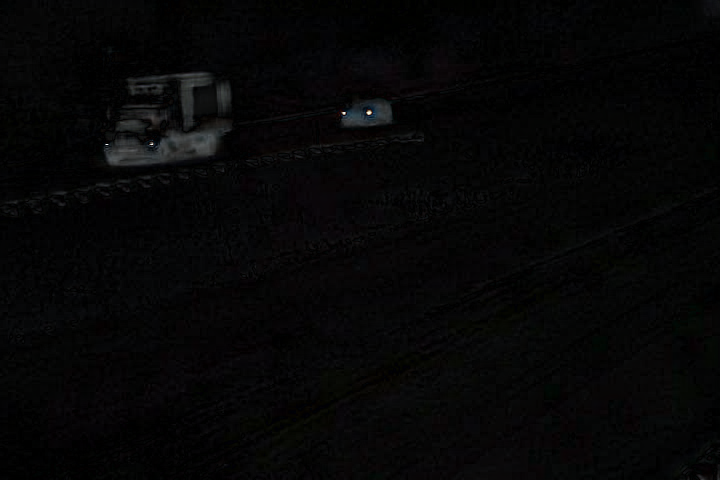}&
      \includegraphics[width=2.8cm, height=2.8cm]{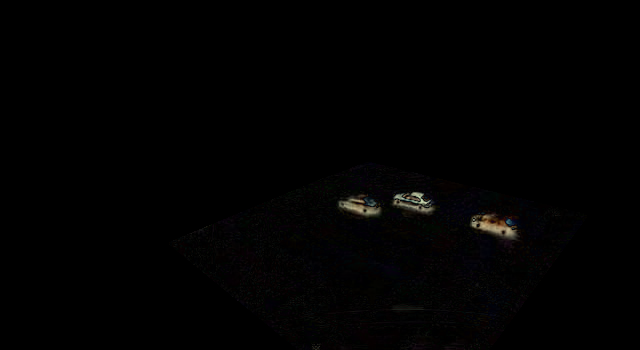}&
      \includegraphics[width=2.8cm, height=2.8cm]{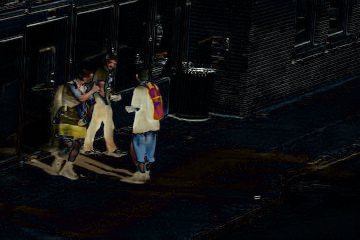}\\
      & (b) & \\
\end{tabular}}
%%%%%%%%%%%%%%%%%% Groundtruth block %%%%%%%%%%%%%%%%%%%%%%
\hspace*{.05cm}
\centerline{
\begin{tabular}{rccc}
      \includegraphics[width=2.8cm, height=2.8cm]{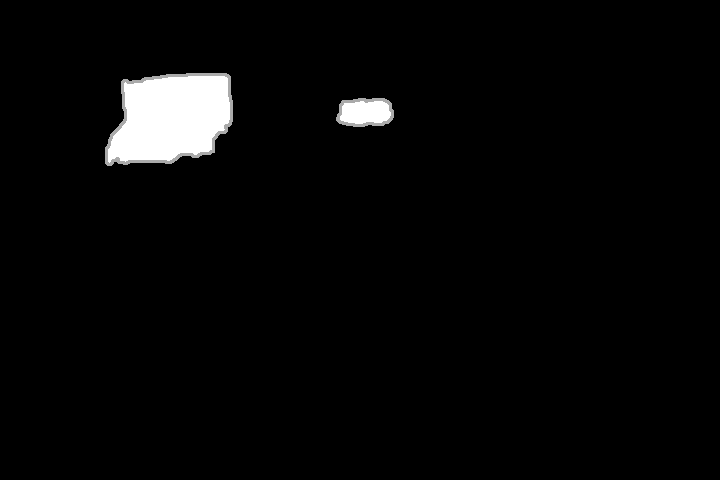}&
      \includegraphics[width=2.8cm, height=2.8cm]{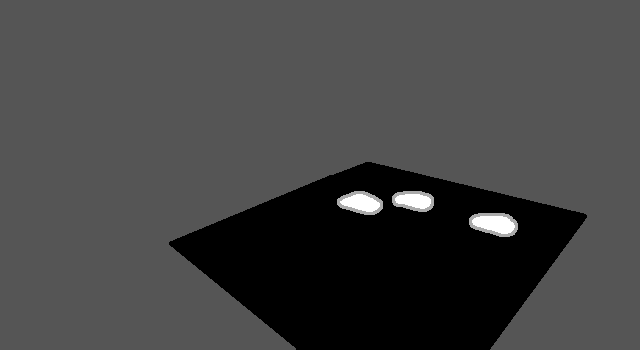}&
      \includegraphics[width=2.8cm, height=2.8cm]{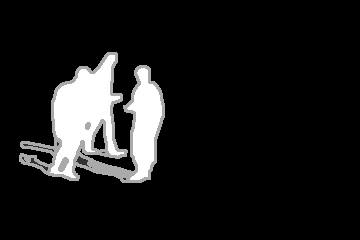}\\
      & (c) & \\
\end{tabular}}
%%%%%%%%%%%%%%%%%% CRCNN_RPM Blocks %%%%%%%%%%%%%%%%%%%%%%
\hspace*{.05cm}
\centerline{
\begin{tabular}{rccc}
      \includegraphics[width=2.8cm, height=2.8cm]{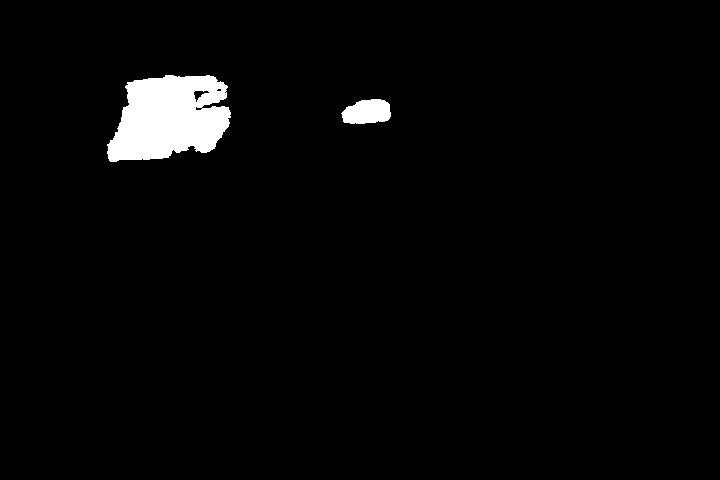}&
      \includegraphics[width=2.8cm, height=2.8cm]{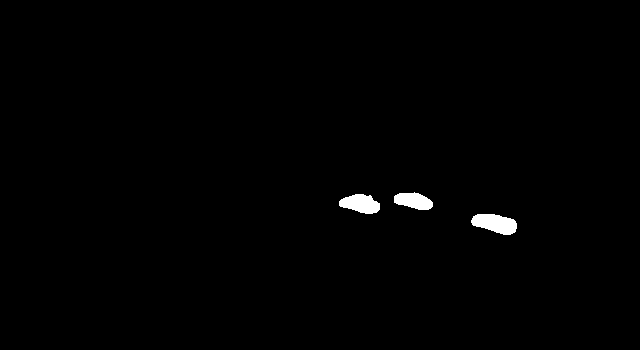}&
      \includegraphics[width=2.8cm, height=2.8cm]{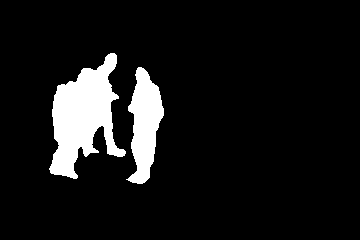}\\
     & (d) & \\
\end{tabular}}
%%%%%%%%%%%%%%%%%% CRCNN Blocks %%%%%%%%%%%%%%%%%%%%%%
\hspace*{.05cm}
\centerline{
\begin{tabular}{rccc}
      \includegraphics[width=2.8cm, height=2.8cm]{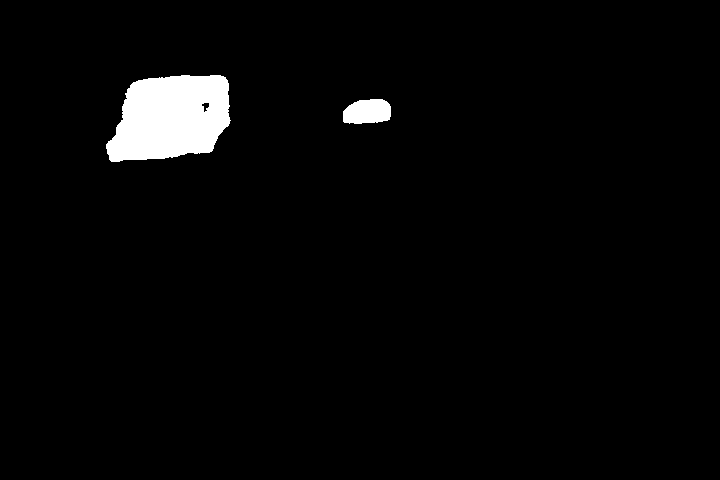}&
      \includegraphics[width=2.8cm, height=2.8cm]{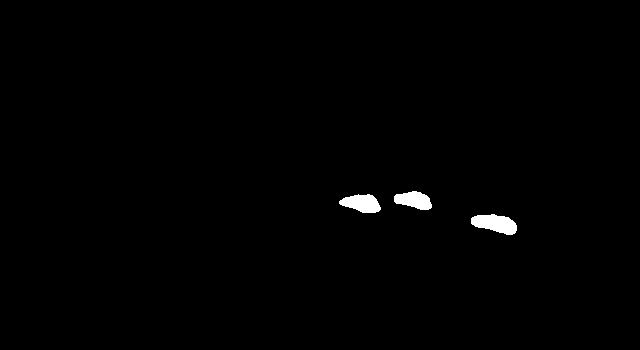}&
      \includegraphics[width=2.8cm, height=2.8cm]{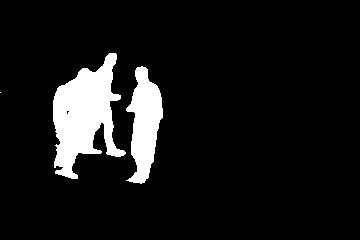}\\
     & (e) & \\
\end{tabular}}
%%%%%%%%%%%%%%%%%% Cascade Blocks %%%%%%%%%%%%%%%%%%%%%%

\hspace*{.05cm}
\centerline{
\begin{tabular}{rccc}
      \includegraphics[width=2.8cm, height=2.8cm]{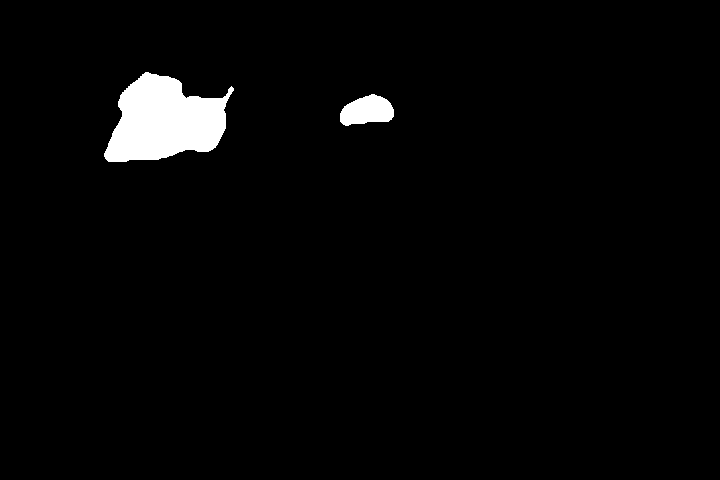}&
      \includegraphics[width=2.8cm, height=2.8cm]{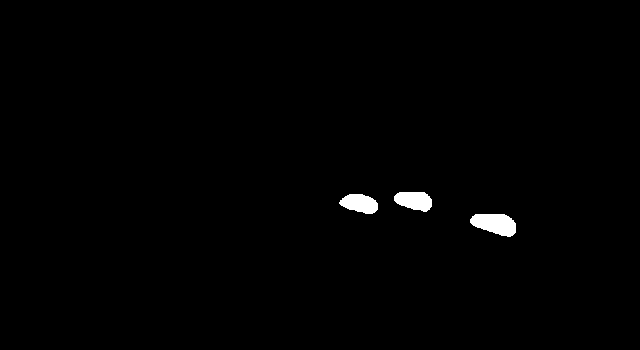}&
      \includegraphics[width=2.8cm, height=2.8cm]{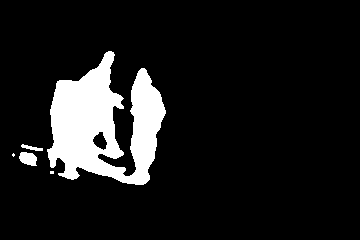}\\
     & (f) & \\
\end{tabular}}

%%%%%%%%%%%%%%%%%% Segnet_V2 Blocks %%%%%%%%%%%%%%%%%%%%%%
\hspace*{.05cm}
\centerline{
\begin{tabular}{rccc}
      \includegraphics[width=2.8cm, height=2.8cm]{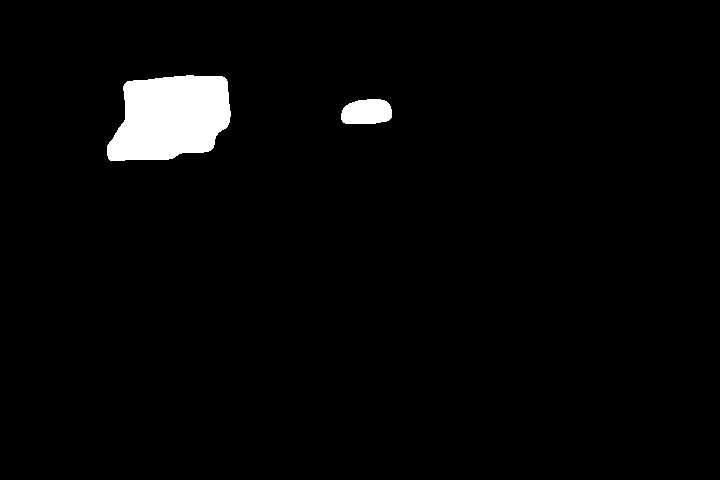}&
      \includegraphics[width=2.8cm, height=2.8cm]{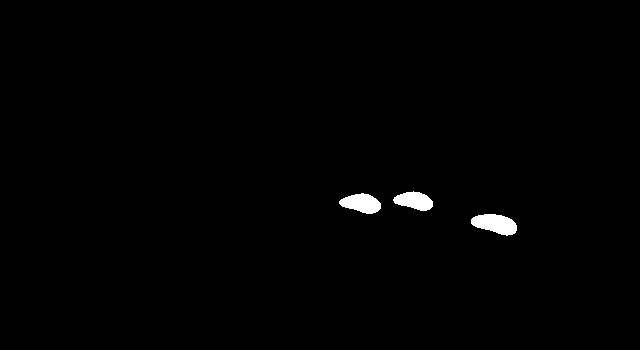}&
      \includegraphics[width=2.8cm, height=2.8cm]{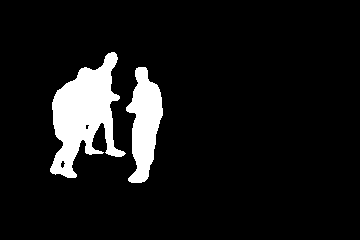}\\
      & (g) & \\
\end{tabular}}
\centering
\caption{Qualitative results considering the categories ``Bad Weather", ``Low Framerate", and ``Shadow" from CD2014 dataset: (a) input RGB frame, (b) MCRCNN residual maps, (c) ground-truth detection masks, results concerning (d) proposed MCRCNN, (e) CRCNN, (f) Cascade and, (g) FgSegNet\_v2.}
\label{f.cd2014_segmentation}
\end{figure}

%%%%%%%%%%%%%%%%%%%%%%%PetrobrasROUTES figures%%%%%%%%%%%%%%%%%%%%%%%%%%%%%%

\section{Conclusion}\label{s.conclusion}

In this work, we proposed a  novel  Cascade  Residual  Convolutional Neural Network that integrates a multiscale processing strategy (through a developed residual processing module) with a learning-based segmentation mechanism in an attempt to solve the scene change detection problems. Regarding tests conducted over CD2014 dataset, the proposed MCRCNN model achieved results close to the state-of-the-art change detection techniques. The proposal was capable of overcoming three supervised learning-based change detection methods and three other non-learning based ones. Even so the MCRCNN did not overcome the FgSegNet\_v2, FgSegNet\_S, and FgSegNet\_M techniques regarding CD2014 dataset, it proven to be much more compact, i.e., around $8\times$ smaller than the best scored FgSegNet\_v2 technique in the number of network parameters. Regarding the test conducted over PetrobrasROUTES dataset, the proposed MCRCNN model outperformed the top two state-of-the-art techniques FgSegNet\_v2 and FgSegNet\_S, and also the CRCNN method. Regarding future works, we pretend to focus our investigation in the MCRCNN false negative problem, conducting a more careful analysis of the RPM filter. We also intend to search for other possible ways to improve the residual learning process and also explore different ways of integrating the residual learned map with the second stage MCRCNN segmentation network.

%==========================================
\iffinal
\section*{Acknowledgment}
The authors are grateful to CNPq grants 307066/2017-7 and 427968/2018-6, FAPESP grants 2013/07375-0 and 2014/12236-1, as well as Petrobras grant 2017/00285-6.
\fi

\bibliographystyle{IEEEtran}

%\bibliography{IEEEabrv, refs}
\bibliography{ms}

\end{document}